\documentclass{article}
\usepackage[preprint]{corl_2026}   
\usepackage{amsmath,amssymb,bm}
\usepackage{graphicx,booktabs,xcolor,xspace,placeins,multirow,wrapfig}
\usepackage{tikz}
\usetikzlibrary{arrows.meta,positioning}
\graphicspath{{img/}{human-rli-paper/img/}}

\newif\ifnotes\notesfalse

\definecolor{notesouju}{RGB}{0,86,201}
\definecolor{notecheck}{RGB}{214,110,0}

\definecolor{cstatic}{HTML}{7F7F7F}
\definecolor{cavoid}{HTML}{E07B39}
\definecolor{crecomp}{HTML}{3B7DD8}
\newcommand{\umbra}{\textsc{umbra}\xspace}
\newcommand{\bench}{\textsc{umbra-bench}\xspace}
\newcommand{\Avoid}{\textsc{Avoid}\xspace}
\newcommand{\Recomp}{\textsc{Recompose}\xspace}
\newcommand{\Static}{\textsc{Static}\xspace}
\newcommand{\Tfree}{T_{\mathrm{free}}}

\title{Sharing the Spotlight: Evaluating Robot Shadow Art\\ in a Shared Human-Robot Environment}

\author{Soujanya C Aryal \quad Simin Liu \quad Ivy Xiao He \quad Adriana Schulz \quad Stefanie Tellex\\
Brown University}

\begin{document}
\maketitle

\begin{abstract}
Robot performance is often evaluated by task success in a fixed environment, even when robots must adapt their behavior around people. We present \umbra, a multi-robot shadow-art system that uses staged, image-space-decomposed optimization to arrange robotic arms into target silhouettes, and a wall camera to adapt those compositions when a person's shadow enters the projection. On an empty wall, \umbra achieves $0.77$ mean IoU across 562 silhouettes using the best result from ten optimization seeds per target, and 34 study participants recognized $85.8\%$ of its simulated and $68.8\%$ of its physical shadows. We then use \umbra to study three aspects of behavior in a shared visual environment: accommodating the person, preserving the task, and verifying predicted outcomes. In simulated intrusions, a reaction that explicitly preserves the target retains more of it than one prioritizing clearance and motion ($0.84$ vs.\ $0.76$), with slightly higher residual overlap. Under simulated model mismatch, predicted clearance often disagrees with the resulting projection. In 11 physical trials with a fixed cardboard-hand stand-in, wall-camera verification and retry let both reactions meet an image-space clearance threshold in the visible margin around the stand-in. These results show why task retention, visual interference, and agreement between predicted and observed outcomes should be reported together.
\end{abstract}


\begin{figure}[!h]
  \centering
  \includegraphics[width=0.5\linewidth]{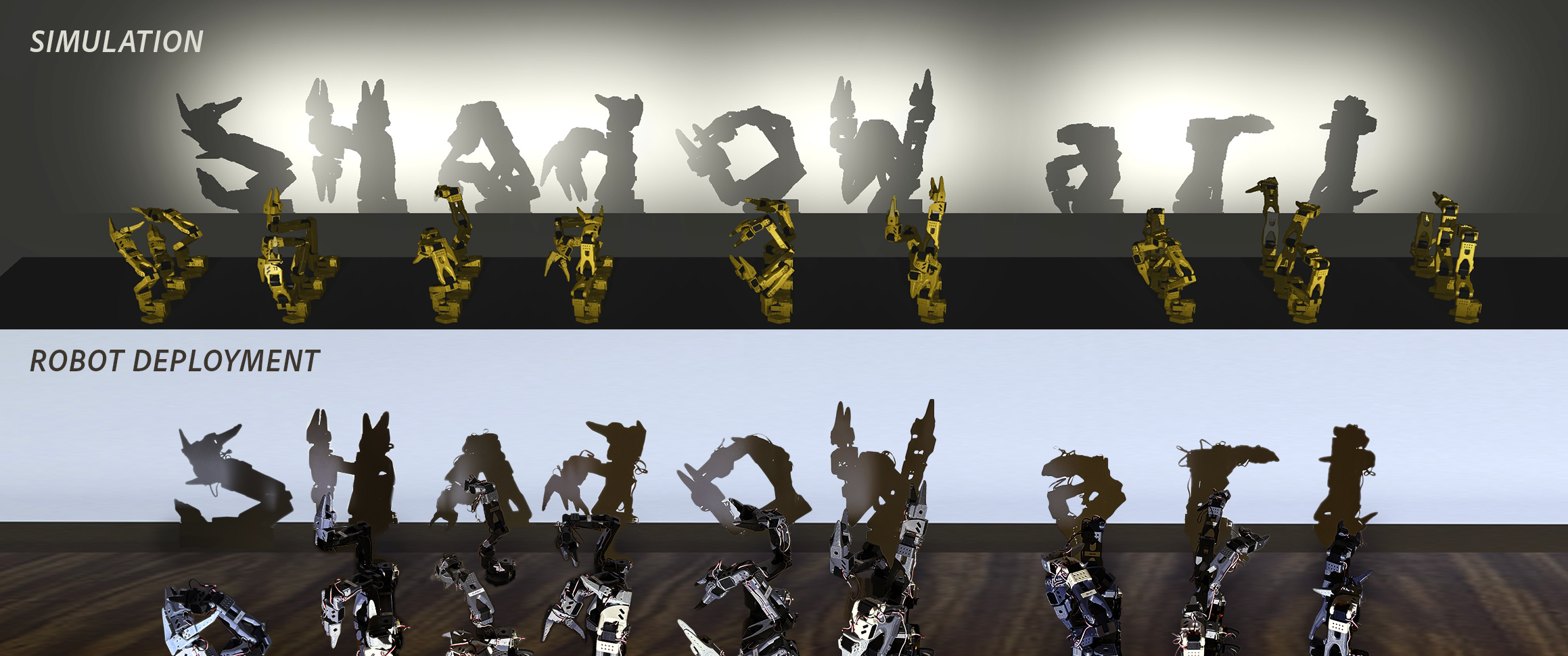}\\[1pt]
  {\small\textbf{(a)} Empty wall: simulated and physical shadows.}\\[3pt]
  \includegraphics[width=0.53\linewidth]{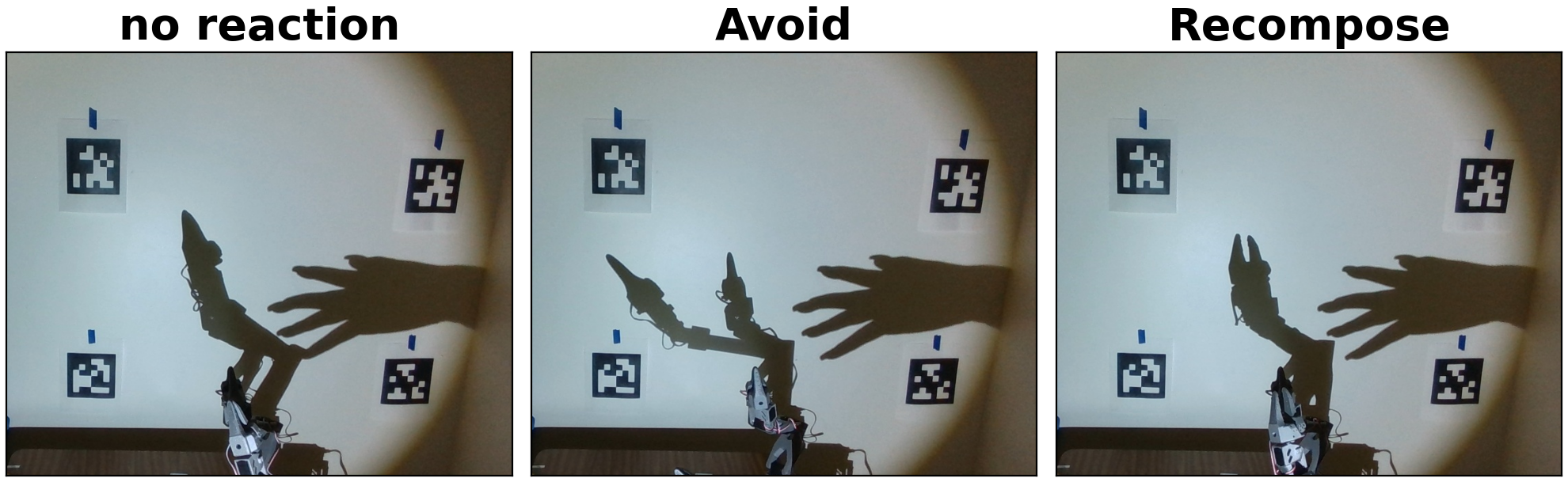}\\[1pt]
  {\small\textbf{(b)} Shared scene: a cardboard stand-in hand's shadow enters the letter \texttt{V} on the physical rig.}
  \caption{\textbf{The evaluation problem.} (a) On an empty wall, success is one IoU score. (b) When a hand's shadow enters, the static pose keeps the letter but covers $6.3\%$ of the visible margin outside the hand; both reactions clear that margin, while \Recomp keeps more of the remaining letter than \Avoid (kept $0.43$ vs.\ $0.27$; static $0.51$). One task score cannot tell these outcomes apart.}
  \label{fig:teaser}
\end{figure}

\section{Introduction}
\label{sec:intro}

Robots are often evaluated by how well they complete a task under fixed conditions. Robots deployed around people, however, must also accommodate changes in their surroundings without needlessly abandoning their tasks. A robot may preserve task performance while interfering with a person, or avoid interference by sacrificing the task altogether, and task success alone cannot distinguish these outcomes. This motivates evaluation methods that measure not only task completion, but also how a robot's behavior affects people sharing its environment, and whether the observed outcome matches the planned one.

Robot shadow art~\cite{mitra2009shadow,won2016shadow,huang2017shadow,yasumoto2019shadow,yan2026robotlearningcommunicateprojected} makes this concrete. We present \umbra, a multi-robot system for composing and adapting shadow art with robotic arms. Given a target silhouette, \umbra searches for joint configurations whose combined shadows reproduce the target on a wall. Its staged optimization decomposes the target across the arms, so the same method composes with different fleet sizes. When a person's shadow enters the projection, \umbra reconfigures the arms to reduce interference while trying to preserve the artwork, and a wall camera observes the result and triggers replanning when the predicted clearance is not achieved.

\textbf{Sharing the wall.} A person's shadow occupies part of the same visual space, and the robots must decide how much of the target to give up. We evaluate three requirements: \textbf{make room} for the person, \textbf{keep the picture} by preserving the task, and \textbf{check the wall} rather than trust a prediction of the outcome. Each captures a different failure: a robot that holds its pose preserves the target while overlapping the person, one that folds its arms away clears the person but abandons the target, and a planner may predict clearance that the physical projection does not provide. An empty-wall IoU sees none of these.

\textbf{Contributions.} First, we present \umbra, a multi-robot shadow-art system that composes target silhouettes through staged optimization and adapts its compositions to human-shadow intrusions using camera-based feedback. Second, we formulate and compare two reactive strategies, \Avoid and \Recomp, that balance image-space interference, task retention and motion. Third, we evaluate \umbra on an empty-wall benchmark, a shadow-recognition study, simulated intrusions, model mismatch and physical hardware (Table~\ref{tab:protocol}), showing why task performance, interference and observed-outcome verification should be considered together in shared environments.

\section{Related Work}
\label{sec:related}

\textbf{Shadow art.} Computational shadow art optimizes static geometry~\cite{mitra2009shadow,sadekar2022shadow}, virtual characters~\cite{won2016shadow} or hand poses~\cite{xu2025hand}. Robots have cast shadows from a pose library~\cite{huang2017shadow}, with a purpose-built 39-DoF caster~\cite{yasumoto2019shadow}, and with a dexterous hand and a learned shadow model~\cite{yan2026robotlearningcommunicateprojected}. Interactive installations have incorporated viewers' silhouettes into visual compositions since \textsc{Videoplace}~\cite{krueger1985videoplace,zhang2025ephemeralshadowhyperrealbeings}. Our focus is different: we use shadow art to evaluate how a robot balances interference with a person's shadow against preservation of its original target, and whether its predicted clearance agrees with the observed projection.

\textbf{Human-aware motion and its evaluation.} Human-aware planners keep robots out of people's space with proxemic costs~\cite{hall1966hidden,sisbot2007human,kruse2013human}, motion prediction~\cite{mainprice2013human} and reactive fields~\cite{khatib1986real}, and safety work measures separation and contact~\cite{haddadin2009requirements,lasota2017survey}. HRI has long argued for task-specific metrics beyond completion~\cite{steinfeld2006common}; we measure task retention and visual interference on a shared wall. Our keep-out margin is defined in the projected image rather than physical space: a pose far from the person can still cover their shadow. Outcome feedback is long established in visual servoing~\cite{hutchinson1996tutorial}, and domain randomization trains against model error offline~\cite{tobin2017domain}; our retry is a simple online form of outcome verification.

\section{\umbra}
\label{sec:umbra}
\label{sec:interact}

\begin{figure}[!htbp]
  \centering
  \includegraphics[width=\linewidth]{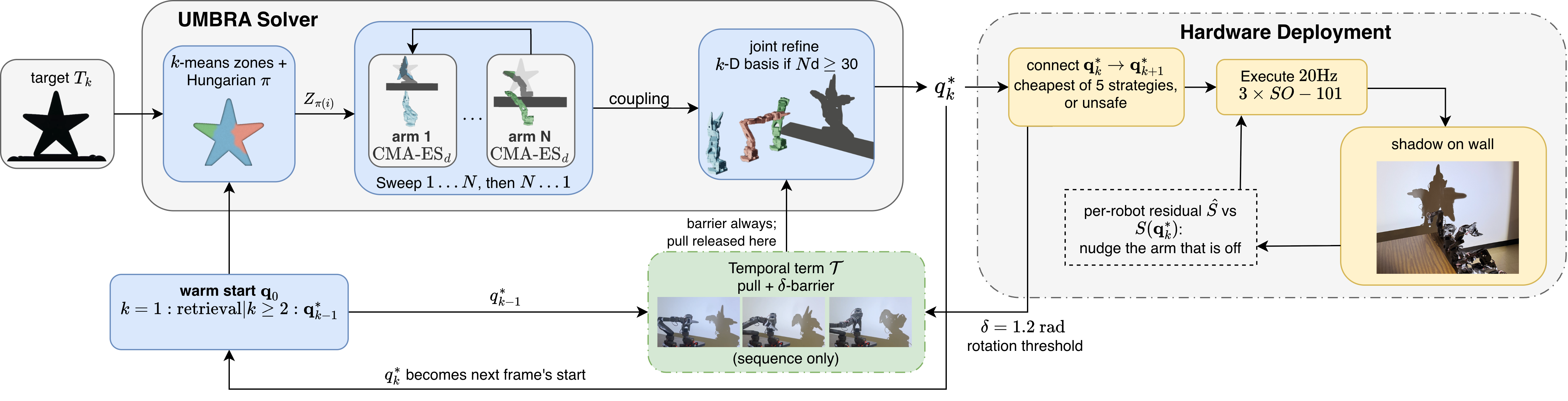}
  \caption{\textbf{\umbra composition.} Target zones, CMA-ES sweeps and joint refinement cast the empty-wall shadow. The diagram also includes sequence-generation components from the empty-wall system; these are not used in the shared-wall experiments.}
  \label{fig:compose-system}
  \vspace{-8pt}
\end{figure}

\umbra is a multi-robot shadow-art system built around one shared optimizer: it \emph{composes} a target on an empty wall (Fig.~\ref{fig:compose-system}), \emph{reacts} when a person's shadow enters (Fig.~\ref{fig:avoidrecomp}), and \emph{verifies} each move. The wall camera provides perception and outcome feedback; the controller executes the resulting arm configurations.

\subsection{Compose: casting a target on an empty wall}

$N$ arms with joints $\bm{q}$ stand between a point light and a wall. Their shadow $S(\bm{q})\in\{0,1\}^{H\times W}$ is the pixelwise union of each arm's shadow. Three properties shape the method. Arm shadows overlap, so an arm's contribution cannot be scored independently of the other arms. $S$ is piecewise constant in $\bm{q}$, so it has no useful gradient. And a single search over every arm's joints grows expensive as arms are added, which motivates splitting the image among the arms and searching in stages. Given a target silhouette $T$, \umbra solves
\begin{equation}
\label{eq:static}
\min_{\bm{q}\in\mathcal{Q}}\ \mathcal{L}_0(\bm{q}) = \alpha\bigl(1-\mathrm{IoU}(S,T)\bigr) + \beta\,\Phi_T(S) + \mathcal{P}_{\mathrm{floor}} + \mathcal{P}_{\mathrm{col}},
\end{equation}
where $\Phi_T$ is the mean distance of shadow pixels from $T$, which provides a search signal when the shadows do not overlap the target, $\mathcal{P}_{\mathrm{floor}}$ penalizes insufficient clearance above the floor/table, and $\mathcal{P}_{\mathrm{col}}$ penalizes inter-arm and self-collision. $\mathcal{Q}$ contains the joint limits. Shadows are rendered in MuJoCo~\cite{todorov2012mujoco}. We cluster target foreground pixels into $N$ zones with $k$-means~\cite{lloyd1982least}, weight boundary pixels threefold, then assign zones to arms using sampled arm--zone overlap costs and the Hungarian algorithm~\cite{kuhn1955hungarian}. Each arm is searched with CMA-ES~\cite{hansen2016cma} while the others are frozen, forward then backward through the fleet, followed by joint refinement. For composition, we retrieve five past targets by cosine similarity of downsampled masks, render their poses against the new target, and start from the best; reactive solves instead start at the current pose.

\subsection{React: re-solving around a person}
\label{sec:react}

\begin{figure}[!htbp]
  \centering
  \resizebox{\linewidth}{!}{\begin{tikzpicture}[
    box/.style={draw=black!60,rounded corners=2pt,align=center,font=\scriptsize,minimum width=22mm,minimum height=9mm,inner sep=2pt,fill=#1},
    box/.default=white,
    flow/.style={-{Latex[length=1.8mm]},thin},
    note/.style={font=\scriptsize,fill=white,inner sep=1pt}
  ]
    \node[box=gray!8] (target) at (0,0) {target $T$};
    \node[box=gray!8] (camera) at (0,-1.6) {wall camera};
    \node[box=gray!8] (mask) at (3,-1.6) {hand mask $S_H$};
    \node[box=blue!7,minimum width=32mm] (compose) at (5.8,0) {compose: $\min\mathcal{L}_0$};
    \node[box=blue!7,minimum width=32mm] (react) at (5.8,-1.6) {react: \Avoid / \Recomp};
    \node[box=gray!8] (arms) at (9.1,-0.8) {execute arm poses};
    \node[box=orange!12] (check) at (9.1,-2.4) {check visible margin};
    \draw[flow] (target) -- (compose);
    \draw[flow] (camera) -- (mask);
    \draw[flow] (mask) -- (react);
    \draw[flow] (compose.east) -| (arms.north);
    \draw[flow] (react.east) -| (arms.south);
    \draw[flow] (arms) -- (check);
    \draw[flow] (check.west) -| node[note,near start,below]{widen and retry} (react.south);
  \end{tikzpicture}}
  \caption{\textbf{Reactive wall loop.} The camera supplies a hand mask; the composer re-solves from the current pose with a keep-out region. After execution, a wall check triggers a wider-margin retry if needed.}
  \label{fig:avoidrecomp}
  \vspace{-8pt}
\end{figure}
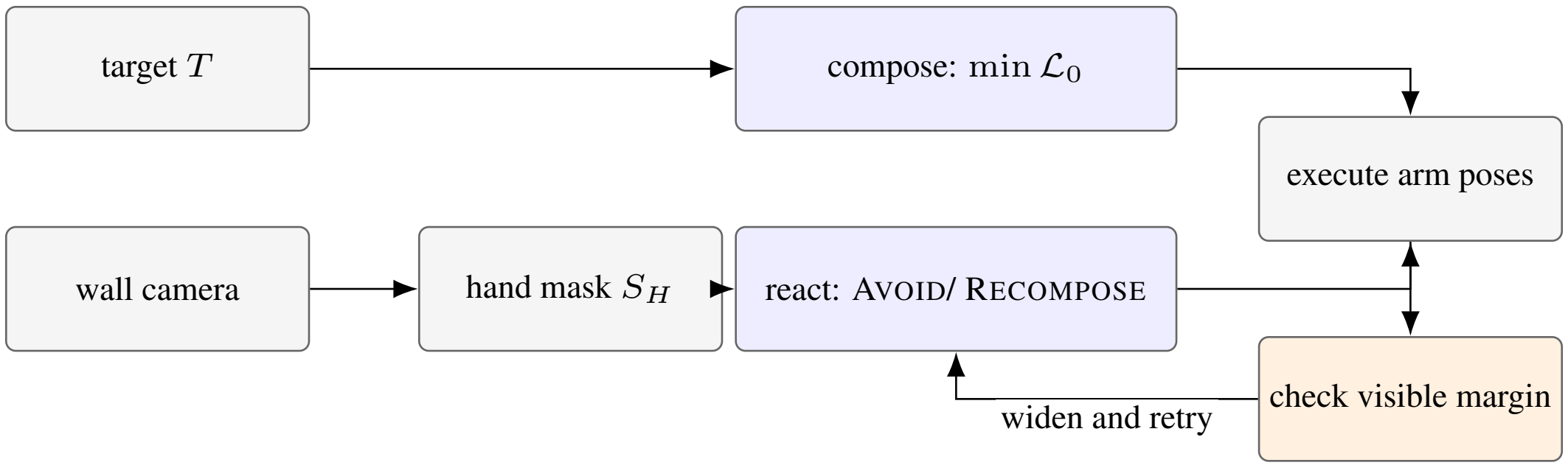

\umbra treats a person's shadow as a changing constraint on the composition problem. Rather than composing a new scene, it starts from the current configuration and re-solves the same objective around the region the person occupies, so the composition optimizer also drives reactive behavior, and the objective decides how much of the artwork is preserved. Let the person's shadow be $S_H$, and let $D_r(S_H)$ be $S_H$ dilated by $r$ pixels (the \emph{margin}), and $\Tfree=T\wedge\neg D_r(S_H)$ the portion of the target outside the person's margin. From the current pose $\bm{q}_c$, \umbra minimizes Eq.~\ref{eq:static} with $T$ replaced by $\Tfree$, plus a keep-out and a smoothness term:
\begin{equation}
\label{eq:loss}
\begin{split}
\mathcal{L}(\bm{q}) = {} & \underbrace{\alpha\bigl(1-\mathrm{IoU}(S,\Tfree)\bigr) + \beta\,\Phi_{\Tfree}(S)}_{\text{\Recomp only}}
+ \lambda_h \tfrac{|S\wedge D_r(S_H)|}{|T|} \\
& + \gamma\bigl\|W\odot(\bm{q}-\bm{q}_c)\bigr\|^2 + \mathcal{P}_{\mathrm{floor}} + \mathcal{P}_{\mathrm{col}}.
\end{split}
\end{equation}
The optimization penalty is normalized by target area $|T|$; evaluation violation is normalized by the evaluated margin area. Two reactions differ only in the shape terms (Fig.~\ref{fig:avoidrecomp}). \Avoid omits the shape terms and favors clearance with smaller joint changes. \Recomp keeps them, so it is rewarded for re-forming strokes around the person. ($\alpha{=}1$, $\beta{=}0.5$, $\lambda_h{=}5$, $\gamma{=}0.5$, $r{=}4$\,px at $128^2$; joint changes are normalized by joint range, and $W$ halves the cost of the base pan.) The staged search is the same as in compose mode, with three changes. It starts at $\bm{q}_c$ with a small step. It re-solves only the arms whose shadow touches the margin. It ranks candidates feasibility first, so any pose that clears the margin beats any that does not, and $\bm{q}_c$ is always a candidate.

\subsection{Closing the loop on the wall}

\begin{wrapfigure}{r}{0.36\linewidth}
  \vspace{-12pt}
  \centering
  \includegraphics[width=\linewidth]{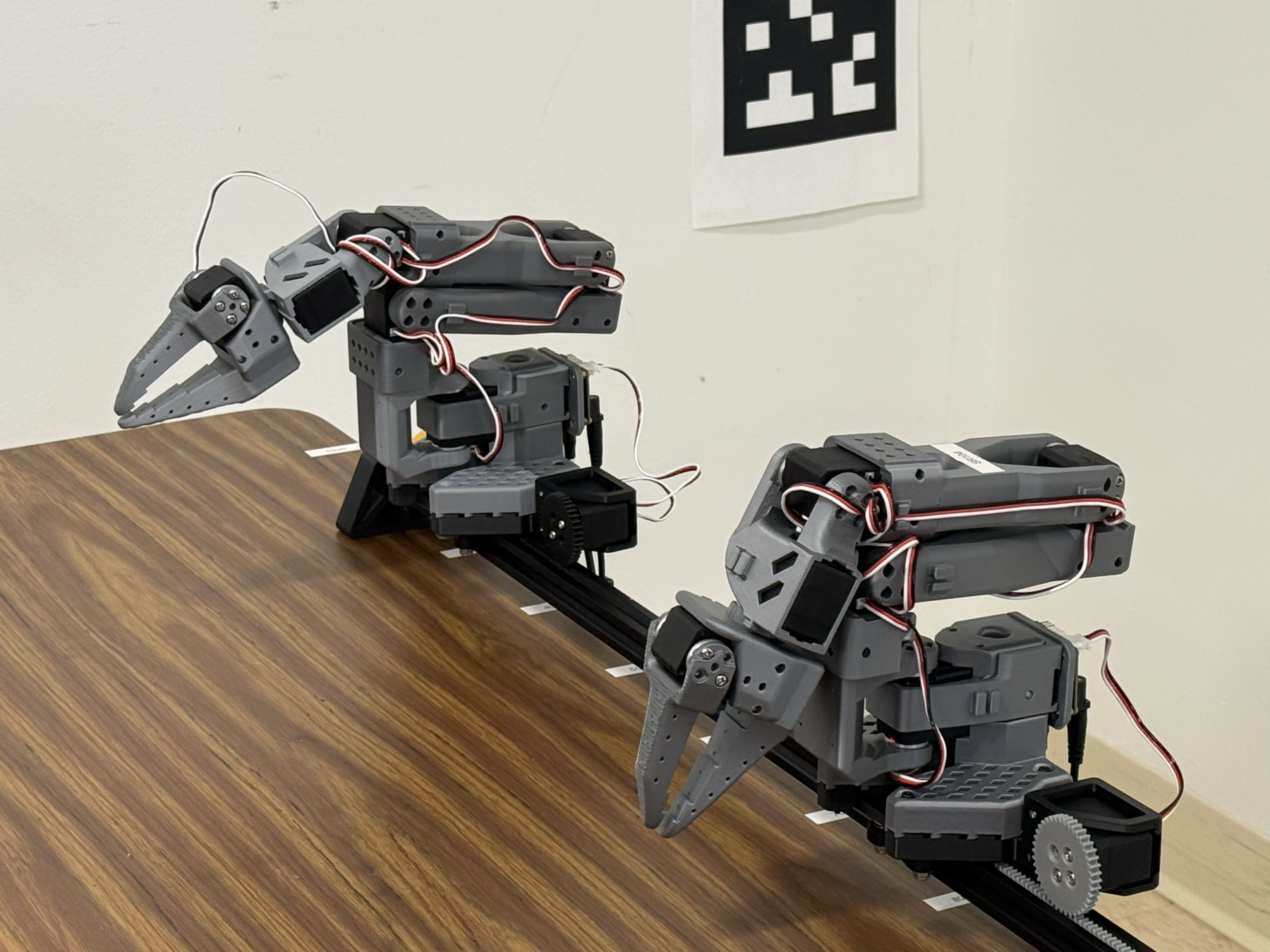}
  \caption{Two SO-101 arms on one rail, facing the tagged wall.}
  \label{fig:rig}
  \vspace{-12pt}
\end{wrapfigure}

A 15\,fps RealSense D435 watches the wall and runs the perception pipeline of App.~Fig.~\ref{fig:pipeline}; two SO-101 arms~\cite{cadene2024lerobot} ride on carriages of one 1\,m rail between a lamp and the wall (Fig.~\ref{fig:rig}). The four tags give the camera-to-wall homography. Starting from tape-measured positions, we then jointly fit five rig parameters (lamp position, wall distance, base height) and an offset for each of the ten arm joints, so that renders match rectified photos of 15 poses; this raises render--photo IoU from $0.12$ to $0.59$, and on poses outside the fit set agreement is $0.44$--$0.65$. Because the predicted shadow can differ from the physical projection, \umbra separates \emph{planning} (in the calibrated model), \emph{execution} (moving the arms) and \emph{verification} (on the wall). After each move, the wall camera measures the margin; if the observed shadow still violates it, the system widens the margin by 3\,px and re-plans, at most twice.

\section{Evaluation}
\label{sec:exp}
\begin{table}[!tb]
\centering\small
\setlength{\tabcolsep}{4pt}
\begin{tabular}{@{}llll@{}}
\toprule
Requirement & Failure exposed & Measure & Where \\ \midrule
Make room & static overlap with person & violation $v$; \% clear & sim; rig \\
Keep the picture & clearance drops the target & retention $\rho$; kept & sim; rig \\
Check the wall & false nominal clearance & false-clear $f$; retry & mismatch; rig \\
Response latency & no intrusion-response timing & re-solve; end-to-end & rig (App.~\ref{app:timing}) \\ \bottomrule
\end{tabular}
\caption{\textbf{Protocol.} Fixed-scene playback does not measure intrusion response.}
\label{tab:protocol}
\vspace{-8pt}
\end{table}

We evaluate the reactions in simulation, under model mismatch, and on hardware (Table~\ref{tab:protocol}).

\textbf{Metrics.} Let $D=D_r(S_H)$ be the dilated person-shadow margin and $R=D\setminus S_H$ its visible ring. For robot shadow $S$ and region $A\in\{D,R\}$, margin violation is $v_A=|S\wedge A|/|A|$. A reaction is clear at $v_A\le\epsilon=0.02$, an operational image-space threshold rather than a physical-safety criterion.
\begin{center}
\small
\setlength{\tabcolsep}{7pt}
\begin{tabular}{@{}lll@{}}
\toprule
Experiment & Predicted $v$ region & Evaluated $v$ region \\ \midrule
Simulated intrusions & full margin $D$ & full margin $D$ \\
Model-mismatch replay & full margin $D$ & visible ring $R$ \\
Hardware stand-in & visible ring $R$ & visible ring $R$ \\ \bottomrule
\end{tabular}
\end{center}
\noindent On hardware, binary wall masks cannot attribute arm-shadow pixels hidden inside $S_H$; $S$ is the observed wall shadow when scoring $R$. In simulation, $S$ is the separately rendered arm mask. \emph{Keep the picture}: simulated intrusions use retention $\rho=\mathrm{IoU}(S,T)/\mathrm{IoU}(S_0,T)$ relative to the pre-intrusion shadow $S_0$; hardware uses \emph{kept} $=\mathrm{IoU}(S_{\mathrm{wall}}\wedge\neg D,\,T\wedge\neg D)$, masking both observed shadow and target outside the margin. \emph{Check the wall} is measured by the false-clear rate
\begin{equation}
\label{eq:metrics}
f=\frac{\#\{i:\ v_{\mathrm{model},i}\le\epsilon\ \wedge\ v_{\mathrm{eval},i}>\epsilon\}}{N_{\text{first solves}}},
\end{equation}
The false-clear rate counts predicted-clear first solves that violate the evaluated threshold, divided by all first solves. Replay compares a full-margin prediction with a perturbed-model ring outcome; hardware compares predicted and observed ring outcomes. We also report clearance after retry.

\begin{figure}[!tb]
\centering
\begin{minipage}[c]{0.53\linewidth}\centering
\setlength{\tabcolsep}{0.5pt}\renewcommand{\arraystretch}{0.9}
\begin{tabular}{cccc}
\includegraphics[width=0.215\linewidth]{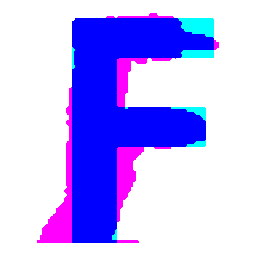} & \includegraphics[width=0.215\linewidth]{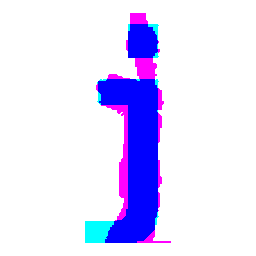} & \includegraphics[width=0.215\linewidth]{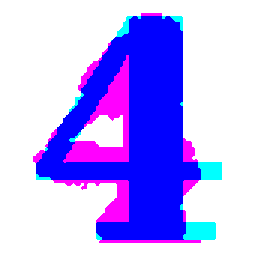} & \includegraphics[width=0.215\linewidth]{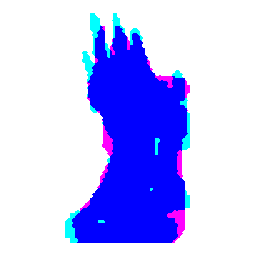} \\[-2pt]
{\scriptsize Upper} & {\scriptsize Lower} & {\scriptsize Digits} & {\scriptsize Teleop} \\[2pt]
\includegraphics[width=0.215\linewidth]{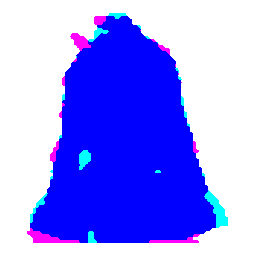} & \includegraphics[width=0.215\linewidth]{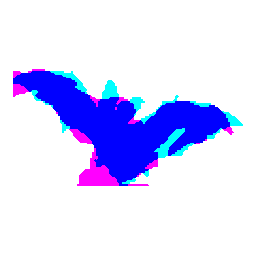} & \includegraphics[width=0.215\linewidth]{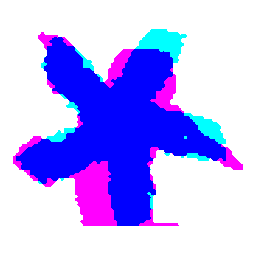} & \includegraphics[width=0.215\linewidth]{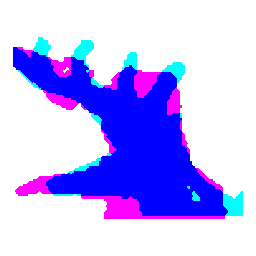} \\[-2pt]
{\scriptsize Objects} & {\scriptsize Animals} & {\scriptsize Abstract} & {\scriptsize Hand} \\[2pt]
\end{tabular}\\[1pt]
{\small\textbf{(a)} \bench: target (cyan) vs.\ \umbra's shadow (magenta); overlap in blue.}
\end{minipage}\hfill
\begin{minipage}[c]{0.45\linewidth}\centering
\includegraphics[width=\linewidth]{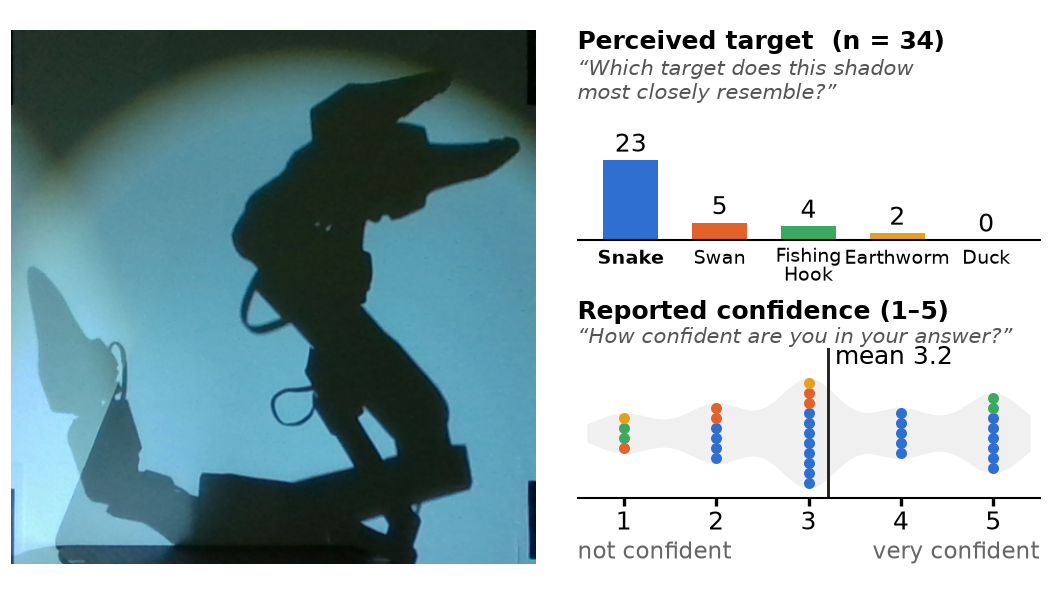}\\[1pt]
{\small\textbf{(b)} User study item: a physically cast ``snake''.}
\end{minipage}
\caption{\textbf{Empty-wall baseline.} (a) One example per \bench subset; across all 562 targets \umbra reaches $0.771$ mean IoU. (b) In a five-way forced-choice study, 34 participants named this physical shadow correctly $68\%$ of the time (mean confidence $3.2/5$; chance $20\%$).}
\label{fig:baseline}
\vspace{-6pt}
\end{figure}

\subsection{Baseline: composing on an empty wall}
\label{sec:baseline}
We first establish that \umbra composes recognizable targets on an empty wall, the reference condition for the cost of accommodating a person. On \bench, 562 silhouettes of letters, digits, objects~\cite{mpeg7}, animals, abstract shapes, hand shadows~\cite{raiyan2025hasperimagerepositoryhand} and teleoperated robot shadows, compose mode with three fixed-base arms reaches $0.771$ mean IoU (best of ten seeds).

Decomposition also reduces the number of renders required for composition as the fleet grows. With five arms on ten probe glyphs, staged search exceeds the best accuracy of a monolithic CMA-ES over all joints using approximately $26\%$ as many renders (each candidate pose costs one render, so renders measure search cost). Reactive re-solves reuse the same staged search. In a five-way forced-choice user study, 34 participants recognized $85.8\%$ of 25 simulated and $68.8\%$ of 5 physically cast shadows (chance $20\%$; Fig.~\ref{fig:baseline}). The physical recognition stimuli came from a three-arm rig (Fig.~\ref{fig:teaser}a) with empty-wall, open-loop playback; all shared-wall experiments below use the two-arm rail rig. The recognition study concerns empty-wall shadows only; it does not measure how people perceive the reactions.

\subsection{Make room and keep the picture: simulated intrusions}
\label{sec:e1}

\textbf{Setup.} We use the calibrated model of our two-arm rig. Twenty \bench targets from six subsets (letters, digits, objects, animals, abstract shapes and hand shadows) are solved for an empty wall. A scripted forearm, with or without a hand, then enters from the left, right or top to $25\%$ or $50\%$ depth ($d$), giving 12 intrusions per target. Each target--intrusion cell has one static episode and three seeds for each reaction (seven episodes per cell; 1680 total). We average reaction seeds within each of 240 cells, then report means $\pm$ s.e.\ across cells.

\begin{figure}[!tb]
  \centering
  \includegraphics[width=0.84\linewidth]{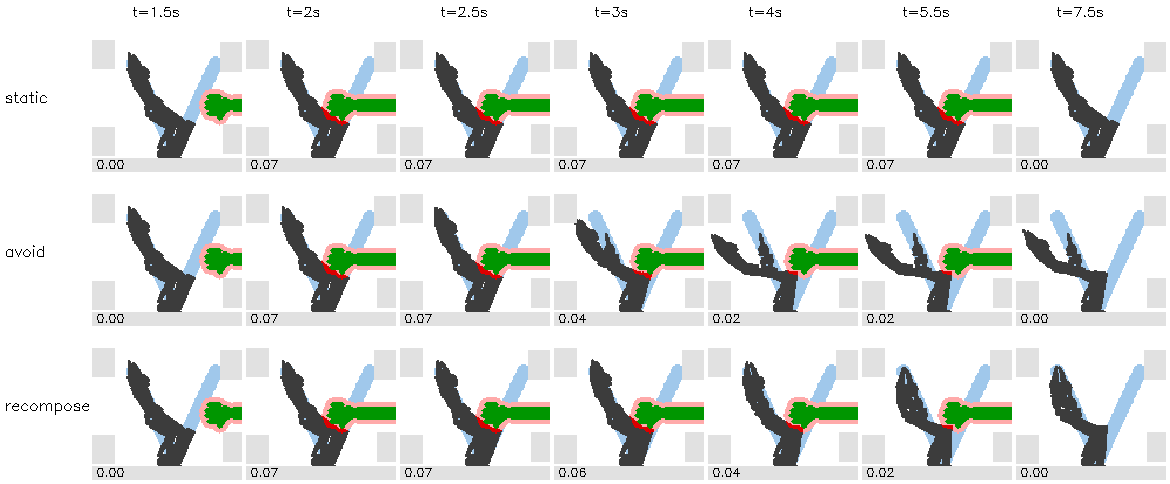}
  \caption{\textbf{Simulated intrusion} (letter \texttt{V}; numbers are violation $v$). As the person's shadow (green) enters and leaves, \Static overlaps it, \Avoid swings an arm aside, and \Recomp re-forms the stroke around it.}
  \label{fig:simfilm}
  \vspace{-6pt}
\end{figure}

\begin{table}[!tb]
\centering\small
\begin{tabular}{@{}lccccc@{}}
\toprule
 & Violation $v\downarrow$ & Retention $\rho\uparrow$ & $\rho$ at $d{=}0.5$ & Joint motion (rad) & Solve (s) \\ \midrule
\Static & $0.163\pm0.009$ & 1.000 & 1.000 & 0 & -- \\
\Avoid  & $\mathbf{0.020}\pm0.001$ & $0.755\pm0.011$ & 0.665 & 2.03 & 0.30 \\
\Recomp & $0.026\pm0.002$ & $\mathbf{0.839}\pm0.008$ & \textbf{0.766} & 2.62 & 0.32 \\ \bottomrule
\end{tabular}
\caption{\textbf{Simulated intrusions} (1680 episodes; mean $\pm$ s.e.\ over 240 target$\times$intrusion cells, seeds averaged within a cell). Solve is the mean of episode-wise maximum re-solve times (one desktop GPU).}
\label{tab:e1}
\vspace{-8pt}
\end{table}

\textbf{Results.} The static baseline preserves the original target by definition, but produces substantial margin violation ($0.163$; Table~\ref{tab:e1}). Both reactions reduce violation substantially. \Avoid achieves lower mean violation than \Recomp ($0.020$ vs.\ $0.026$), while \Recomp retains more of the target ($0.839$ vs.\ $0.755$; paired gain $+0.084\pm0.004$, larger in 204 of 240 cells) and performs better on deeper intrusions. \Recomp also requires more joint motion ($2.62$ vs.\ $2.03$\,rad). These results show a trade-off between reducing image-space interference and preserving the task, rather than an unconditional improvement in both metrics; Fig.~\ref{fig:simfilm} shows one episode. The slowest re-solve per episode has a median of $0.25$\,s, and on average 1.1 of the 2 arms move.

\textbf{Motion regularization (App.~Table~\ref{tab:e2}).} Removing motion regularization reduces \Avoid's margin violation ($0.018\to0.008$), but increases joint motion ($1.87\to8.32$\,rad) and reduces target retention ($0.761\to0.671$). This illustrates why clearance should be evaluated alongside artwork preservation and movement.

\subsection{Check the wall: simulated model mismatch}
\label{sec:e3}

\begin{figure}[!tb]
  \centering
  \includegraphics[width=0.64\linewidth]{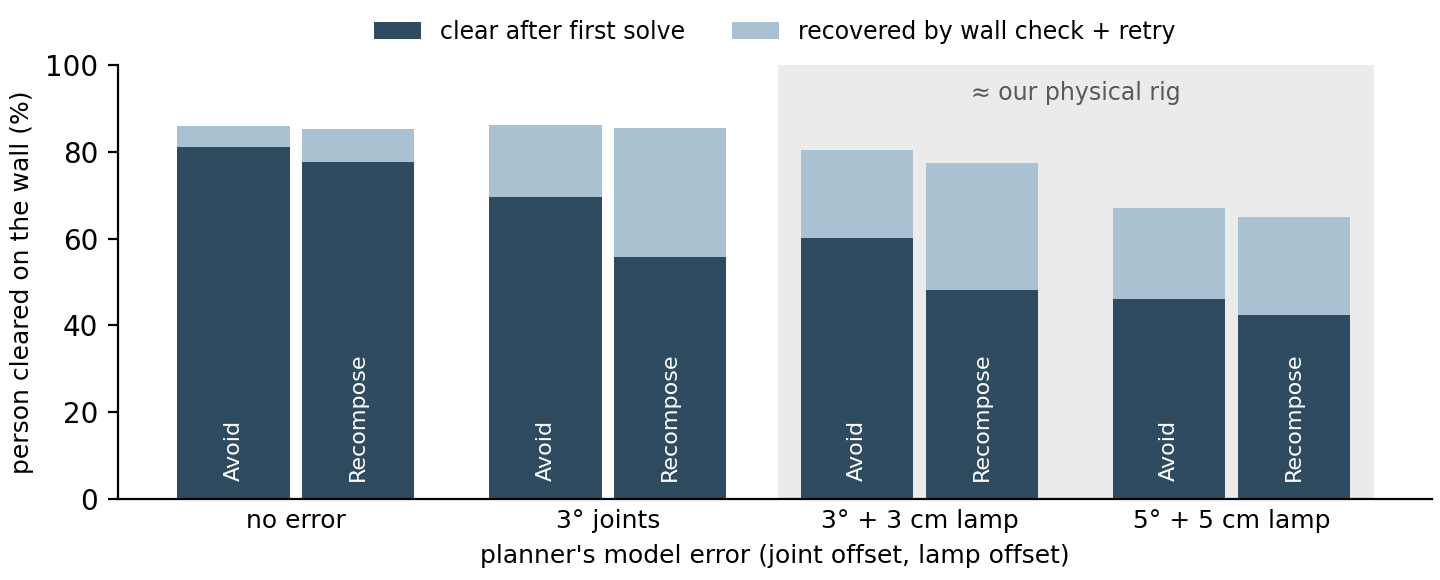}
  \caption{\textbf{Clearance under model mismatch} (558 contact cases). Dark bars show first-solve clearance in the perturbed evaluation model; light bars show additional cases cleared after verification and retry. Section~\ref{sec:e3} separately reports failures that the nominal model incorrectly predicted as clear.}
  \label{fig:mismatch}
  \vspace{-8pt}
\end{figure}

We plan in the nominal model and evaluate with random joint/lamp offsets. From 80 recorded hand-shadow frames, four letters and four error levels (none; $3^\circ$ joints; $3^\circ$/$3$\,cm; $5^\circ$/$5$\,cm), we retain only frame--letter--level cases whose static pose violates the perturbed-model visible-ring threshold: 143, 145, 133 and 137 cases (558 total; 415 perturbed). \emph{Perturbed-model kept} is $\mathrm{IoU}(S_{\mathrm{eval}}\wedge\neg D_4(S_H),\,T\wedge\neg D_4(S_H))$, using the same 4\,px evaluation margin before and after retry. At $3^\circ$/$3$\,cm error, $60\%$ (\Avoid) and $48\%$ (\Recomp) of first solves clear the perturbed model (Fig.~\ref{fig:mismatch}). At hardware-like levels (shadow IoU $0.55$/$0.43$ vs.\ rig $0.44$--$0.65$), false-clear rates are $25$--$43\%$ ($17$--$30\%$ with joint error alone). Even with no injected error, $5$--$8\%$ of first solves are predicted clear over the full margin but violate the visible-ring threshold, reflecting the difference between these evaluation definitions. One or two retries raise clearance to $80\%$/$77\%$ with little retention loss (\Recomp kept $0.30\to0.29$ at $3^\circ$/$3$\,cm); after retry, clearance differs by at most three percentage points at each level, and \Recomp keeps more in $72\%$ of perturbed cases.

\subsection{All three on hardware: stand-in trials}
\label{sec:rig}

\begin{figure}[!tb]
  \centering
  \includegraphics[width=0.56\linewidth]{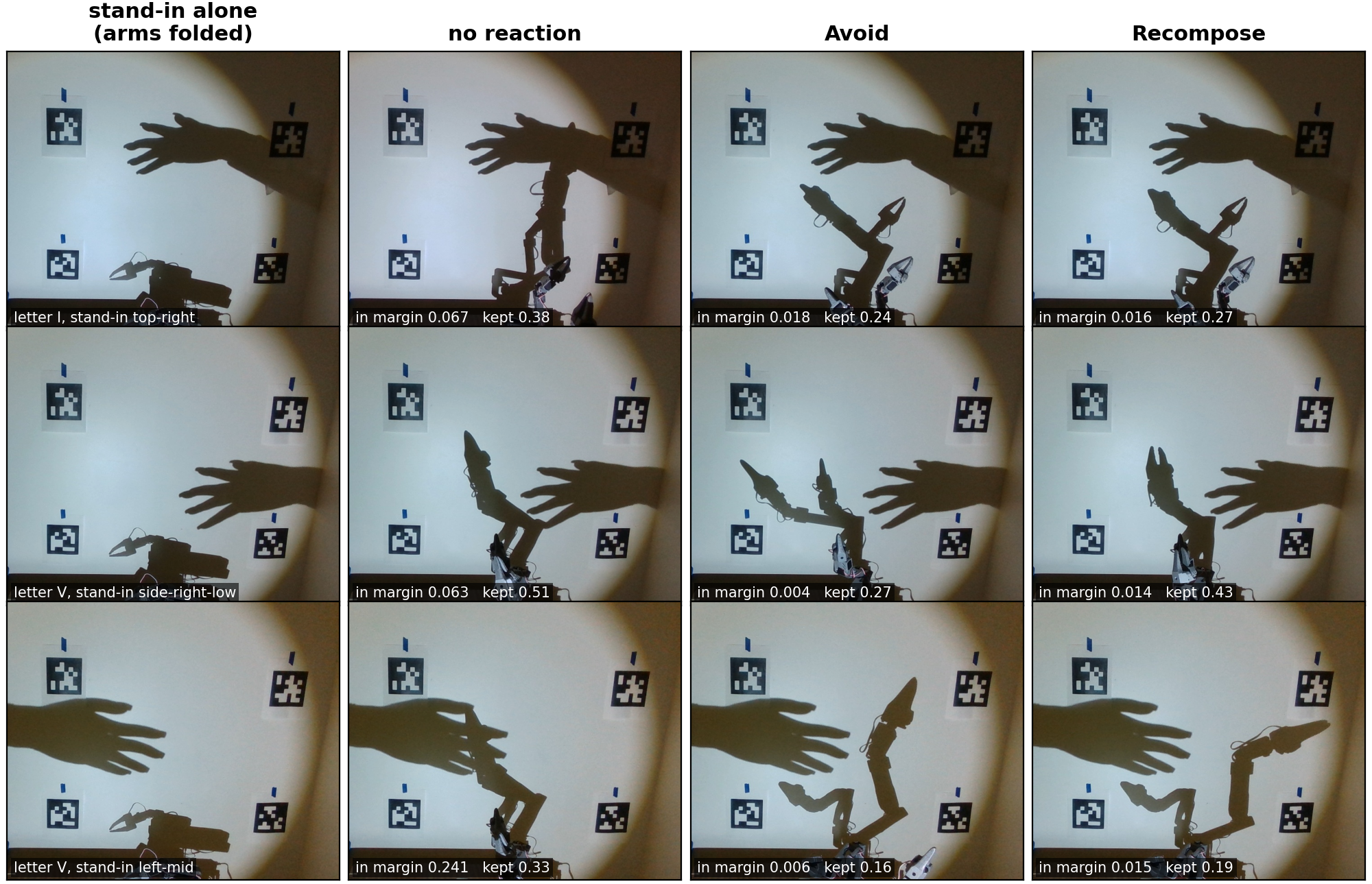}
  \caption{\textbf{Three physical trials.} Left: stand-in shadow $S_H$ with arms folded. Other columns: static pose and both reactions. \Recomp keeps more of each letter than \Avoid (trial k11: $0.43$ vs.\ $0.27$ kept; Table~\ref{tab:rig}).}
  \label{fig:hwgrid}
  \vspace{-8pt}
\end{figure}

\textbf{Stand-in hand.} We use a rigid cardboard hand to hold the intrusion fixed across \Static, \Avoid and \Recomp. An arms-folded photo minus the folded-arm reference yields $S_H$ (Fig.~\ref{fig:hwgrid}). Later binary wall masks recover arm-shadow pixels only outside $S_H$; hardware \emph{kept} therefore masks the entire dilated margin, while clearance and retry verify the threshold only in the surrounding visible ring. From k9 onward, we also photographed the folded-arm wall at each trial's end: retained k10--k15 start/end masks agree at IoU $\ge0.98$; k9's end photo was too bright to score. Each trial poses a letter (\texttt{I}, \texttt{T}, \texttt{V}, \texttt{X}) and runs both reactions with retry. Of 17 trials over four placements, 4 had no letter--margin contact. We excluded two more: one with an unexplained change, likely a moved stand-in before end-of-trial checks, and one with the room light on, which invalidates the dark-room mask protocol. Eleven trials remain, including two (k7, k15) with briefly brighter ambient light but consistent stand-in shadows.

\begin{figure}[!tb]
  \centering
  \includegraphics[width=\linewidth]{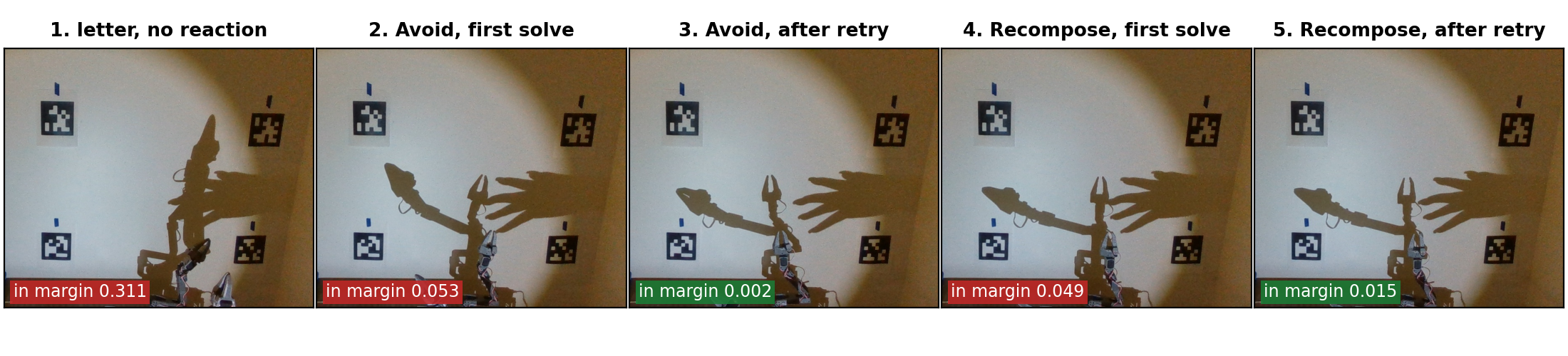}
  \caption{\textbf{Check the wall, on hardware} (trial k9, letter \texttt{I}). Both first solves were clear in the planner's model but not on the wall (red). Re-solving with a 3\,px wider margin cleared them (green).}
  \label{fig:recovery}
  \vspace{-6pt}
\end{figure}

\begin{table}[!tb]
\centering\small
\begin{tabular}{@{}lcccc@{}}
\toprule
 & Violation on wall & Clear, 1st solve & Clear, after retry & Kept \\ \midrule
\Static & $0.174\pm0.113$ & 0/11 & -- & $0.43\pm0.12$ \\
\Avoid  & $0.010\to\mathbf{0.006}$ & 10/11 & \textbf{11/11} & $0.20\pm0.11$ \\
\Recomp & $0.016\to0.012$ & 9/11 & \textbf{11/11} & $\mathbf{0.26}\pm0.12$ \\ \bottomrule
\end{tabular}
\caption{\textbf{Physical rig, 11 stand-in trials} (mean $\pm$ sample s.d.; arrows: first solve $\to$ after retry). Clearance and violation are measured in the visible ring outside the stand-in shadow; kept is IoU after masking both the observed shadow and target outside the dilated margin. The planner predicted $0.001$/$0.007$ ring violation for its first solves. \Recomp kept more than \Avoid in 11/11 trials.}
\label{tab:rig}
\vspace{-8pt}
\end{table}

\textbf{Results.} Hardware \emph{kept} measures target overlap outside the margin, not retention relative to the pre-intrusion shadow. Without a reaction, the letter covers $17\%$ of the visible ring on average and up to $39\%$ (Table~\ref{tab:rig}). \emph{Make room}: after retry, both reactions clear the visible ring in all 11 trials. \emph{Keep the picture}: \Recomp keeps more of the letter than \Avoid in every trial (Fig.~\ref{fig:hwgrid}; per trial in App.~Figs.~\ref{fig:k11} and~\ref{fig:bars}), with slightly higher residual violation ($0.012$ vs.\ $0.006$). \emph{Check the wall}: the planner predicted 21 of 22 first solves clear, the wall showed that 3 of those were not (false-clear rate $f{=}3/22$ over all first solves, or $3/21$ among predicted-clear first solves), and retries recovered all 3 (Fig.~\ref{fig:recovery}). Wall checks thus raised visible-ring clearance from 19/22 to 22/22 reactions.

\subsection{Real-hand replay and latency}
\label{sec:replay}
To complement the stand-in trials with a real hand, we replayed a four-minute recording of a hand entering and holding in the beam against the solver with \emph{simulated} move durations (949 contact frames). The static pose violated the margin in all contact frames. With simulated movement durations of 1.45\,s, the fraction of contact frames with margin violation fell to $58\%$ for \Avoid and $68\%$ for \Recomp; with 0.75\,s moves, it fell to $39\%$ and $45\%$ (App.~\ref{app:timing}). No false detections occurred over 143 empty-wall frames. Live on the rig, median re-solve latency is $0.45$\,s and end-to-end response latency (detection to arms in place) about $1.7$\,s, mostly a 1.2\,s move ramp we chose rather than a servo limit.

\FloatBarrier
\section{Discussion and Limitations}
\label{sec:discussion}

\textbf{Beyond shadows and next steps.} The same evaluation dimensions may matter for projected workspaces, collaborative drawing and robot camera operation, though these domains are untested. A user study could establish which reaction people prefer; a learned shadow model could enable predictive reactions.

\textbf{Limitations.} Real-hand recordings support perception and solver replay, while physical clearance and retry are evaluated with a rigid stand-in on one two-arm rig (four letters, 11 trials). Sim-to-real agreement is $0.44$--$0.65$ IoU, and our model-error study perturbs only joints and lamp position. Perception assumes a dark room, and the camera cannot attribute arm-shadow pixels hidden within the stand-in shadow. Retry widens the margin without correcting the model. Image-space clearance is not physical safety, and these experiments do not establish how people perceive the reactions.

\section{Conclusion}
\umbra composed 562 empty-wall targets at $0.77$ mean IoU (best of ten seeds) and adapted its shadows to intrusions. In simulation, \Recomp preserved more artwork than \Avoid (retention $0.839$ vs.\ $0.755$), with slightly more visual overlap; on hardware, verification and retry cleared the visible ring in all 22 reactions across 11 fixed-stand-in trials. These results motivate reporting retention, interference and observed outcomes together, without treating image-space clearance as physical safety.

\clearpage
\IfFileExists{humanrli.bib}{\bibliography{humanrli}}{\bibliography{human-rli-paper/humanrli}}

\clearpage
\appendix
\renewcommand{\thetable}{A\arabic{table}}\setcounter{table}{0}
\renewcommand{\thefigure}{A\arabic{figure}}\setcounter{figure}{0}
\renewcommand{\theHtable}{A\arabic{table}}
\renewcommand{\theHfigure}{A\arabic{figure}}
\section{Additional results}
\label{app:abl}

\begin{table}[h]
\centering\small
\setlength{\tabcolsep}{5pt}
\begin{tabular}{@{}lccc@{}}
\toprule
Condition & Violation $v\downarrow$ & Retention $\rho\uparrow$ & Joint motion (rad) \\ \midrule
\Avoid, with motion regularization & 0.018 & 0.761 & 1.87 \\
\Avoid, without motion regularization & 0.008 & 0.671 & 8.32 \\
\Recomp, with motion regularization & 0.023 & 0.845 & 2.45 \\
\Recomp, without motion regularization & 0.019 & 0.885 & 6.68 \\ \bottomrule
\end{tabular}
\caption{\textbf{Effect of motion regularization.} All four conditions use the same ablation subset (10 targets $\times$ 12 intrusions $\times$ 3 seeds, $r{=}4$); values need not match the full evaluation in Table~\ref{tab:e1}. S.e.\ $\le0.018$ for $\rho$ and $\le0.003$ for $v$.}
\label{tab:e2}

\end{table}

\begin{figure}[h]
  \centering
  \includegraphics[width=0.9\linewidth]{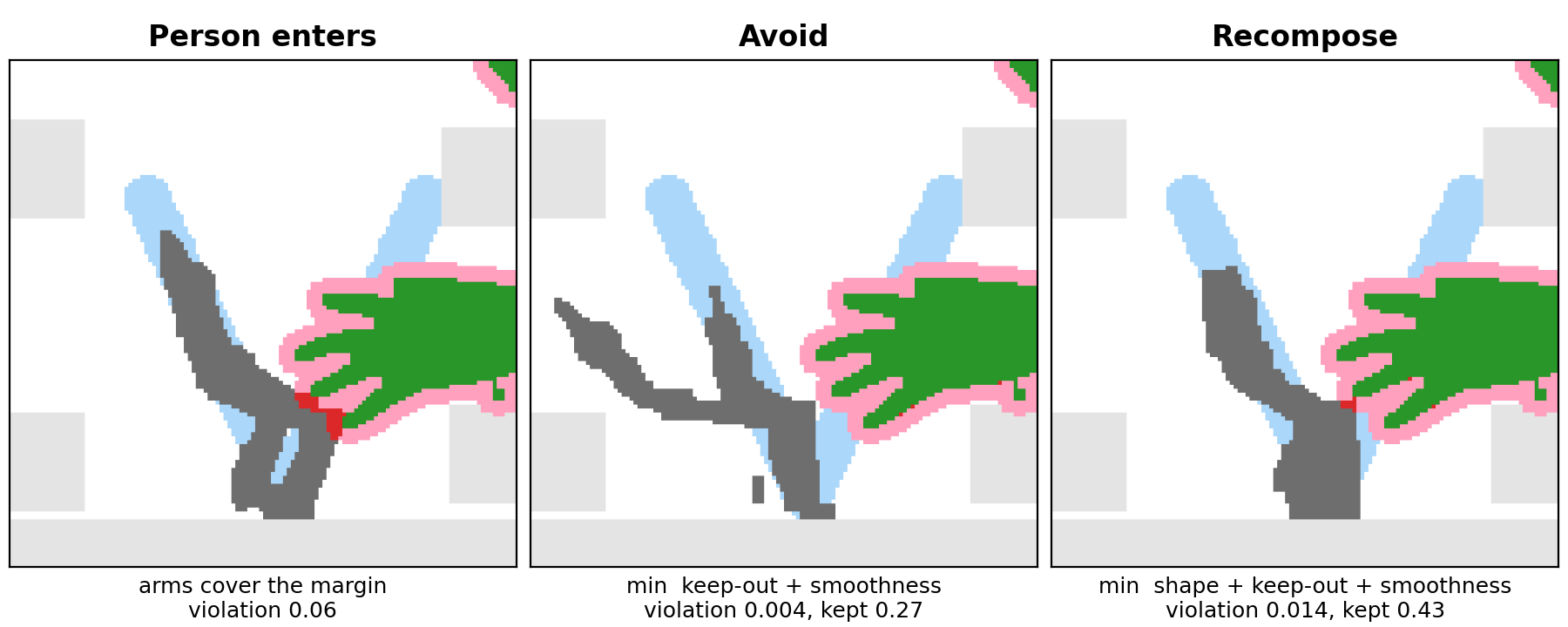}
  \caption{\textbf{Avoid vs.\ Recompose on the wall} (physical trial k11, rectified wall frame: letter in light blue, person's shadow in green, margin in pink, arm shadow in grey, arm shadow inside the margin in red). Both reactions start from the same pose and clear the visible margin. \Avoid minimizes only keep-out and motion, so it swings an arm out of the letter. \Recomp also rewards covering $\Tfree$, so it re-forms the left stroke next to the hand.}
  \label{fig:k11}
  
\end{figure}

\section{Real hands and timing}
\label{app:timing}

\begin{figure}[!tb]
  \centering
  \includegraphics[width=\linewidth]{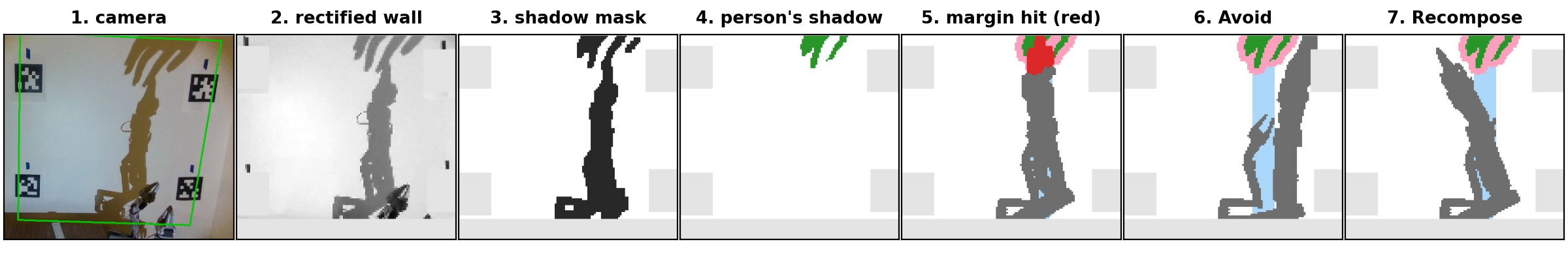}
  \caption{\textbf{Real-hand perception and solver replay.} A recorded frame is rectified and segmented (1--4); panels 5--7 show rendered arm shadows before and after each reaction, not wall-measured outcomes.}
  \label{fig:pipeline}
\end{figure}

\begin{table}[h]
\centering\small
\begin{tabular}{@{}lccc@{}}
\toprule
4-min real-hand replay (949 contact frames) & \shortstack{Contact frames with\\margin violation (\%)} & Mean violation & Kept \\ \midrule
\Static & 100\% & 0.138 & 0.57 \\
\Avoid, 1.45\,s / 0.75\,s moves & 58\% / 39\% & 0.053 / 0.031 & 0.39 / 0.34 \\
\Recomp, 1.45\,s / 0.75\,s moves & 68\% / 45\% & 0.074 / 0.037 & 0.50 / 0.47 \\ \bottomrule
\end{tabular}
\caption{A recorded real hand replayed against the solver with simulated move times. A contact frame counts as a margin violation when $v>\epsilon=0.02$.}
\end{table}

\begin{table}[h]
\centering\small
\begin{tabular}{@{}lr@{}}
\toprule
Stage & Measured on the rig \\ \midrule
Camera & 15.06\,fps, 0 dropped; frame interval p95 69\,ms \\
Rectify + mask / segment person / margin check & 8.4 / 0.2 / 0.07\,ms (p50) \\
Re-solve (live setting / full budget) & 0.45\,s / 1.2--3.7\,s (laptop CPU) \\
Serial read both arms / write one arm & 2.0 / 0.2\,ms \\
Servo 15$^\circ$ step: within $1^\circ$ / final error & 0.35\,s / 0.3--0.5$^\circ$, no overshoot \\
Command $\to$ visible change on wall & 0.14--0.19\,s \\ \midrule
Detection $\to$ arms in place, live (1.2\,s ramp) & $\approx$1.7\,s \\ \bottomrule
\end{tabular}
\caption{System timing on the physical rig. With step moves instead of a ramp, detection to arms-in-place would be about $0.8$\,s (projected).}
\end{table}

\begin{figure}[h]
  \centering
  \includegraphics[width=\linewidth]{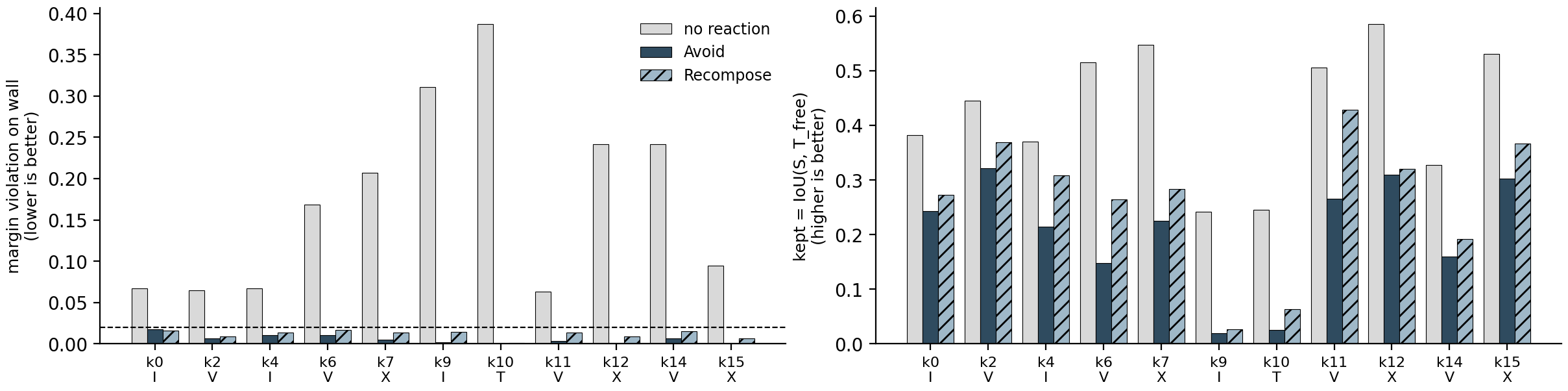}
  \caption{\textbf{Per-trial results for the 11 retained stand-in trials.} Left: margin violation measured on the wall after retry (lower is better; dashed line: threshold $\epsilon{=}0.02$). Right: kept is IoU of the observed shadow and target, both masked outside the dilated margin (higher is better). Light grey: no reaction; dark: \Avoid; hatched: \Recomp.}
  \label{fig:bars}
\end{figure}

\end{document}